\documentclass{article}
\usepackage{spconf,amsmath,amssymb,epsfig}
\usepackage{hyperref}

\title{Field-scale soil moisture estimated from Sentinel-1 SAR data using a knowledge-guided deep learning approach}

\name{
\begin{tabular}{c}Yi Yu$^{a,b,*}$, Patrick Filippi$^{a}$, Thomas F. A. Bishop$^{a}$\end{tabular}\thanks{$^{*}$Correspondence to Yi Yu (yi.yu1@sydney.edu.au)}
}

\address{
$^{a}$The University of Sydney, Sydney, NSW 2006, Australia \\
$^{b}$CSIRO Agriculture and Food, Canberra, ACT 2601, Australia
}

\begin{document}

\maketitle

\begin{abstract}

Soil moisture (SM) estimation from active microwave data remains challenging due to the complex interactions between radar backscatter and surface characteristics. While the water cloud model (WCM) provides a semi-physical approach for understanding these interactions, its empirical component often limits performance across diverse agricultural landscapes. This research presents preliminary efforts for developing a knowledge-guided deep learning approach, which integrates WCM principles into a long short-term memory (LSTM) model, to estimate field SM using Sentinel-1 Synthetic Aperture Radar (SAR) data. Our proposed approach leverages LSTM's capacity to capture spatiotemporal dependencies while maintaining physical consistency through a modified dual-component loss function, including a WCM-based semi-physical component and a boundary condition regularisation. The proposed approach is built upon the soil backscatter coefficients isolated from the total backscatter, together with Landsat-resolution vegetation information and surface characteristics. A four-fold spatial cross-validation was performed against in-situ SM data to assess the model performance. Results showed the proposed approach reduced SM retrieval uncertainties by 0.02 m$^3$/m$^3$ and achieved correlation coefficients (R) of up to 0.64 in areas with varying vegetation cover and surface conditions, demonstrating the potential to address the over-simplification in WCM.

\end{abstract}


\section{Introduction}

Soil moisture (SM) is a critical state variable in land surface hydrology that regulates the exchange of water and heat energy between the land surface and the atmosphere \cite{seneviratne_investigating_2010}. An accurate and high-resolution estimation of SM is crucial as an initial step in agricultural-related and broader environmental studies. Despite its importance, obtaining high-resolution and spatially complete estimates of surface SM remains challenging, particularly at the field scale necessary for agricultural applications \cite{babaeian_ground_2019}. Satellite remote sensing has emerged as a powerful tool for monitoring SM content, offering various approaches to achieve this goal. The most common operational method for retrieving SM utilises the passive microwave electromagnetic spectrum \cite{entekhabi_soil_2010}, which can offer several advantages, including day and night operability, minimal atmospheric interference, and reliability and wide applicability in SM retrievals. However, the trade-off is their coarse spatial resolution, usually in the range of tens of kilometres, which limits their applicability in agricultural field-scale studies. Despite the explorations aimed at deriving high-resolution SM estimates \cite{yu_continental_2021, yu_empirical_2024}, knowledge gaps persist regarding their reliability and uncertainty characterisation across environmental conditions \cite{peng_roadmap_2021, yu_spatial_2025}.

Alternatively, C-band Synthetic Aperture Radar (SAR) sensors present potential for SM retrieval at higher resolution \cite{fan_sentinel-1_2025}. The SAR systems operate by actively transmitting electromagnetic waves and measuring the returned backscatter signal, offering distinct advantages over passive microwave sensors \cite{wagner_temporal_2008}. The strength of SAR lies in its capacity to achieve high spatial resolution, typically ranging from 10 to 30 metres, while maintaining consistent data acquisition capability regardless of atmospheric conditions or solar illumination \cite{torres_gmes_2012}. However, SAR-based SM retrieval also presents inherent signal complexities compared to passive microwave sensors \cite{bauer-marschallinger_toward_2019}, due to the multiplicative nature of radar backscatter interactions with surface characteristics, including vegetation cover, surface roughness and soil texture \cite{oh_empirical_1992}.

The SAR-based SM retrieval predominantly relied on either empirical or data-driven approaches. The empirical and semi-empirical approaches, such as the change detection method \cite{bauer-marschallinger_toward_2019} and the water cloud model (WCM) \cite{ulaby_radar_1996}, attempt to characterise the complex interactions between radar backscatter and vegetation and surface properties \cite{attema_vegetation_1978}. These models typically decompose the total backscatter signal into contributions from soil and vegetation, establishing analytical relationships based on theoretical understanding and experimental observations. However, their performance may be constrained by the simplified assumptions and site-specific parametrisation that may not generalise well across diverse agricultural landscapes \cite{joseph_soil_2008}. In contrast, recent developments in data-driven approaches, particularly deep neural networks, demonstrate potential for field-scale SM estimation by capturing complex, non-linear relationships between SAR observations and predictors. Nonetheless, the purely data-driven models were usually criticised due to the lack of interpretability \cite{singh_piml-sm_2024}.

The integration of semi-physical and data-driven approaches, hence, presents an opportunity to embed the physical insights into the learning capacity of deep neural networks \cite{singh_piml-sm_2024}. By reformulating the semi-empirical relationships within a deep learning framework while preserving physical constraints, such hybrid approaches can potentially overcome the limitations of both paradigms, including both the oversimplification inherent in empirical models and the physical inconsistency risks associated with purely data-driven methods \cite{karpatne_theory-guided_2017}. Therefore, this research presents preliminary efforts to estimate field-scale SM from Sentinel-1 SAR data using a knowledge-guided deep learning approach, to explore the possibility of embedding the knowledge from a WCM into a neural network architecture.

\section{Study area and data}

\subsection{Study area and in-situ data}

\begin{figure}
    \centering
    \includegraphics[width=0.45\textwidth]{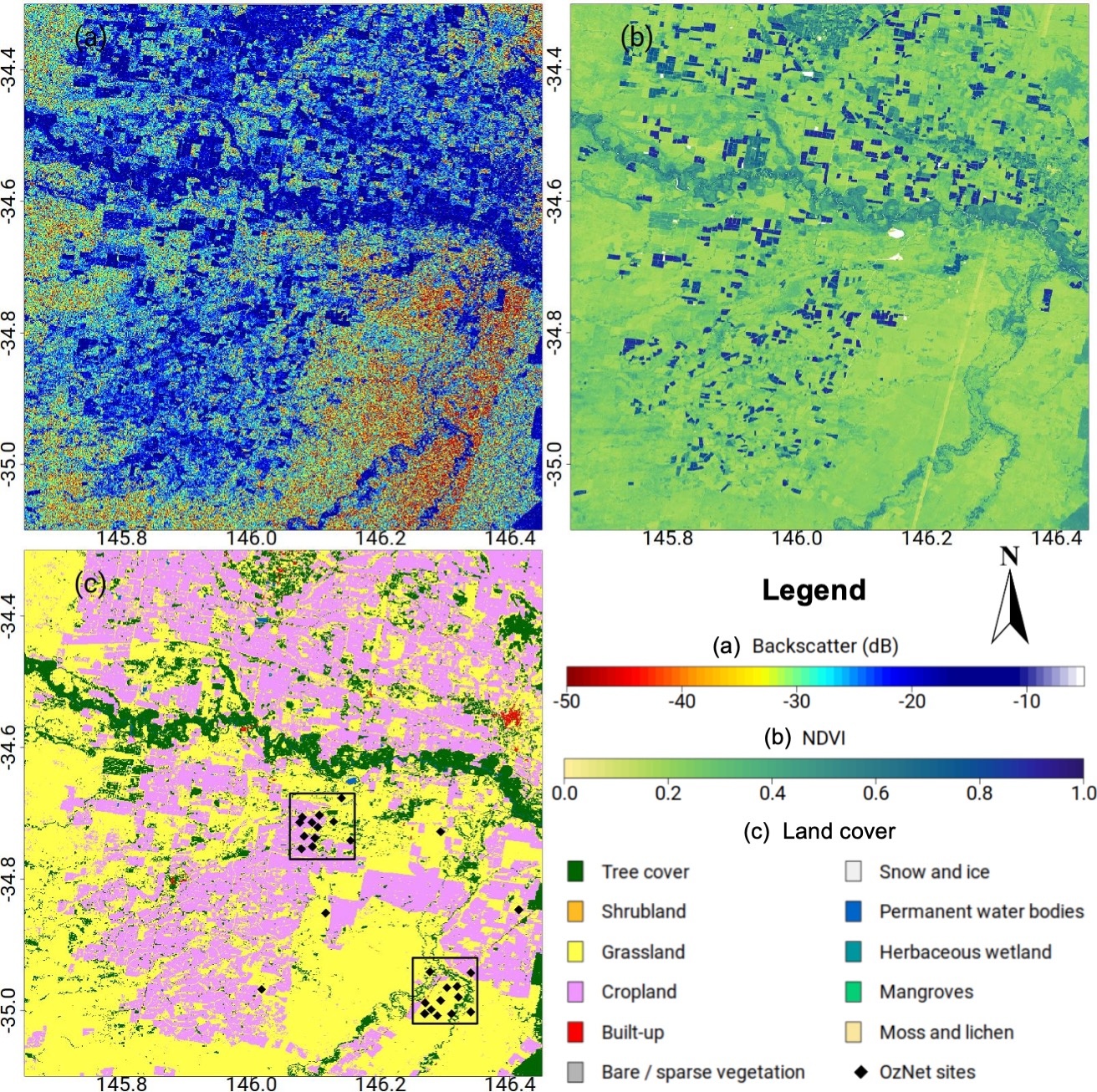}
    \caption{(a) The spatial distribution of Sentinel-1 backscatter coefficient on 01 Feb 2019; (b) the synthesised NDVI derived from the fusion of MODIS and Landsat data on 01 Feb 2019; and (c) the land cover classification based on the ESA WorldCover 10 m v200 dataset \cite{zanaga_esa_2022}.}
    \label{fig1}
\end{figure}

The Yanco agricultural region was chosen as study area for this research, which is situated within the Murrumbidgee Catchment and exemplifies a semi-arid agricultural environment of southeastern Australia. Fig. \ref{fig1} shows an 80 km × 80 km area of (a)  the Sentinel-1 backscatter coefficient data acquired on 01 Feb 2019; (b) the synthesised NDVI data derived from the fusion of MODIS and Landsat surface reflectance coincident with the SAR acquisition time; and (c) the land cover information \cite{zanaga_esa_2022}. The in-situ near-surface SM measurements (0-5 cm) from the OzNet Hydrological Monitoring Network \cite{smith_murrumbidgee_2012} were collected for cross-validation, with sites denoted by black squares in Fig. \ref{fig1} (c). The distribution of OzNet sites was relatively clustered (Fig. \ref{fig1} c). They were then partitioned into four distinct groups to facilitate a four-fold cross-validation, where the detailed information for the folds and clusters can be found in \cite{yu_spatial_2025} (see their Table II).

\subsection{Sentinel-1 SAR data}

The Sentinel-1 mission, operated under the ESA's Copernicus program, comprises twin C-band (5.405 GHz) SAR satellites: Sentinel-1A and -1B \cite{torres_gmes_2012}. The mission offers high-resolution SAR imagery with a spatial resolution of 5 × 20 meters in the Interferometric Wide (IW) swath mode, which represents the primary acquisition mode over land. The constellation achieves a temporal resolution of 6 days at the equator when both satellites are operational, though actual revisit frequency varies by geographical location (e.g., a typical 12-day revisit cycle for eastern Australia).

The Ground Range Detected (GRD) product was collected for this research, which undergoes systematic preprocessing using the Sentinel-1 Toolbox to generate calibrated, ortho-corrected imagery. The preprocessing workflow encompasses thermal noise removal, radiometric calibration, and terrain correction. The final backscatter coefficients are converted to decibel scale through a logarithmic transformation $(10 \times log10(x))$. We specifically focused on the vertical transmit / horizontal receive (VH) polarisation, which has demonstrated sensitivity to SM variations. Each scene includes an additional $angle$ band containing the local incidence angle information derived through interpolation of the geolocation grid points. The Sentinel-1 data was obtained from the Google Earth Engine (GEE) platform \cite{gorelick_google_2017}.

\subsection{Landsat-resolution surface reflectance}

The Landsat 8 mission provides multispectral imagery spanning visible to thermal infrared wavelengths at spatial resolutions ranging from 30 m (visible/near-infrared) to 100 m (thermal), with a 16-day revisit frequency. The Digital Earth Australia (DEA) Nadir BRDF-adjusted Reflectance (NBAR) Landsat-8 collection \cite{li_evaluation_2010}, which implemented a bidirectional reflectance distribution function (BRDF) to ensure consistency with the MODIS NBAR MCD43A4 product \cite{schaaf_first_2002}, was collected for this research. Albedo and Normalised Difference Vegetation Index (NDVI) values were derived from both Landsat and MODIS surface reflectance products. To obtain temporally continuous information at field scale, an unbiased spatiotemporal fusion approach \cite{yu_generating_2023} was then implemented to produce both daily 100 m albedo and NDVI estimates.

\subsection{Auxiliary information}

The auxiliary information of surface characteristics was also collected, including available water capacity, near-surface soil information (clay, sand, silt) and climate grids. A complete list and description of these auxiliaries can be found in \cite{yu_spatial_2025} (see their Table I).

\section{Methods}

\subsection{Water cloud model}

The WCM characterises the radar backscatter interactions between soil and vegetation. The total backscatter coefficient (linear) $\sigma_{\text{obs, linear}}$ is expressed as \cite{xing_retrieval_2025}:

\begin{equation}
    \sigma_{\text{obs, linear}}=\sigma_{\text {veg, linear}}+\gamma^2\times \sigma_{\text {soil, linear}}
    \label{eq1}
\end{equation}

where $\sigma_{\text{veg, linear}}$ and $\sigma_{\text{soil, linear}}$ represent the vegetation contribution (linear) and soil backscatter (linear), respectively; and $\gamma^2$ accounts for the two-way vegetation attenuation. They can be calculated as:

\begin{equation}
    \sigma_{\text{veg, linear}} = A \times \cos \theta \times (1-\gamma^2)
    \label{eq2}
\end{equation}

\begin{equation}
    \sigma_{\text{soil}} = C + D \times \mathrm{SM}
    \label{eq3}
\end{equation}

\begin{equation}
    \gamma^2=e^{-2\tau / \cos \theta}
    \label{eq4}
\end{equation}

\begin{equation}
    \tau=B \times \mathrm{VWC}
    \label{eq5}
\end{equation}

where $A$ is a vegetation backscattering factor; $\theta$ is the local incidence angle (40°); $C$ and $D$ are intercept and slope of a linear regression, respectively, between the soil backscatter (dB) $\sigma_{\text{soil}}$ and SM; $\tau$ is the vegetation optical depth, which can be derived using an empirical vegetation parameter $B$ (0.084 for grassland) and the vegetation water content $\mathrm{VWC}$ (calculated using NDVI). Hence, by reformulating Eq. \ref{eq1} and Eq. \ref{eq3} we can have $\mathrm{SM}_\text{WCM}$ as:

\begin{equation}
    \mathrm{SM}_\text{WCM} = \frac{10 \times \log_{10}((\sigma_{\text {obs, linear}}-\sigma_\text {veg, linear}) / \gamma^2)-C}{D}
    \label{eq6}
\end{equation}

The parameter $A$, $C$, and $D$ can be optimised by minimising the difference between $\mathrm{SM}_\text{WCM}$ and the reference SM (in-situ SM herein). However, the assumption of linear relationships between SM and soil backscatter coefficients requires careful re-consideration, particularly given the complex nature of soil-radar interactions across diverse surface characteristics. A recent investigation has demonstrated the necessity of spatially explicit parameter calibration to account for heterogeneous soil properties and surface conditions \cite{xing_retrieval_2025}. To address this limitation, we propose a knowledge-guided neural network architecture, aiming to capture the inherent nonlinearity of SM-backscatter relationships while maintaining physical consistency with established theoretical frameworks.

\subsection{Knowledge-guided long short-term memory}

The Long Short-Term Memory (LSTM) networks \cite{hochreiter_long_1997} represents a specialised recurrent neural network architecture designed for sequential data analysis. It has demonstrated efficacy in environmental modelling by capturing both short-term fluctuations and long-term patterns in geophysical time series. A typical loss function uses the mean squared error (MSE) between the prediction and the observation:

\begin{equation}
    \mathrm{Loss}_{\text {LSTM}}=\frac{1}{n} \sum_{i=1}^n (\mathrm{SM}_{\text{pred}, i} - \mathrm{SM}_{\text {obs}, i})^2
    \label{eq7}
\end{equation}

where $\mathrm{Loss}_{\text {LSTM}}$ is the loss function of LSTM; $\mathrm{SM}_{\text{pred}, i}$ is the SM prediction at time step $i$; $\mathrm{SM}_{\text {obs}, i}$ is the SM observation at time step $i$. Here we modified it as:

\begin{equation}
    \mathrm{Loss}_{\text {modified}}= \mathrm{Loss}_{\text {soil}} + \mathrm{Loss}_{\text{boundary}}
    \label{eq8}
\end{equation}

\begin{equation}
    \mathrm{Loss}_{\text{soil}}= \frac{1}{n} \sum_{i=1}^n (\mathrm{SM}_{\text{pred, soil}, i} - \mathrm{SM}_{{\text{obs}}, i})^2
    \label{eq9}
\end{equation}

\begin{equation}
\begin{split}
    \mathrm{Loss}_{\text{boundary}}= \lambda \times \frac{1}{n} \sum_{i=1}^n[\max (0,-\mathrm{SM}_{\text{pred, soil}, i})+ \\
    \max (0, \mathrm{SM}_{\text{pred, soil}, i}-1)]
    \label{eq10}
\end{split}
\end{equation}

where $\mathrm{Loss}_{\text{modified}}$ is the modified loss function consisting of the soil component contribution $\mathrm{Loss}_{\text{soil}}$ and physical boundaries $\mathrm{Loss}_{\text{boundary}}$; $\lambda$ is a regularisation factor; $\mathrm{SM}_{\text{pred, soil}}$ is the WCM knowledge-guided SM prediction (using isolated soil backscatter), which is expressed as:

\begin{equation}
    \mathrm{SM}_{\text {pred, soil}} = f_{\text{LSTM}}([{\sigma_{\text{soil}}, \mathbf{v}}]_{t-n:t})
    \label{eq11}
\end{equation}

\begin{equation}
    \sigma_{\text {soil}} = 10 \times \log_{10}\frac{\sigma_\text{obs, linear} - A \times \cos \theta \times\left(1-\gamma^2\right)} {\gamma^2}
    \label{eq12}
\end{equation}

\begin{equation}
    A=e^{\log A}
    \label{eq13}
\end{equation}

where $f_{\text{LSTM}}$ represents the LSTM model that is used to replace the linear regression in Eq. \ref{eq6}; $\mathbf{v}$ represents the vector of auxiliary surface characteristics used in the LSTM regression; $\sigma_{\text{soil}}$ is the soil backscatter (dB) calculated using WCM knowledge, with $A$ being a learnable parameter to be optimised by the LSTM model; $A=e^{\log A}$ constrains $A$ to positive values while allowing unconstrained optimisation.

\section{Results and discussion}

\begin{figure}
    \centering
    \includegraphics[width=0.48\textwidth]{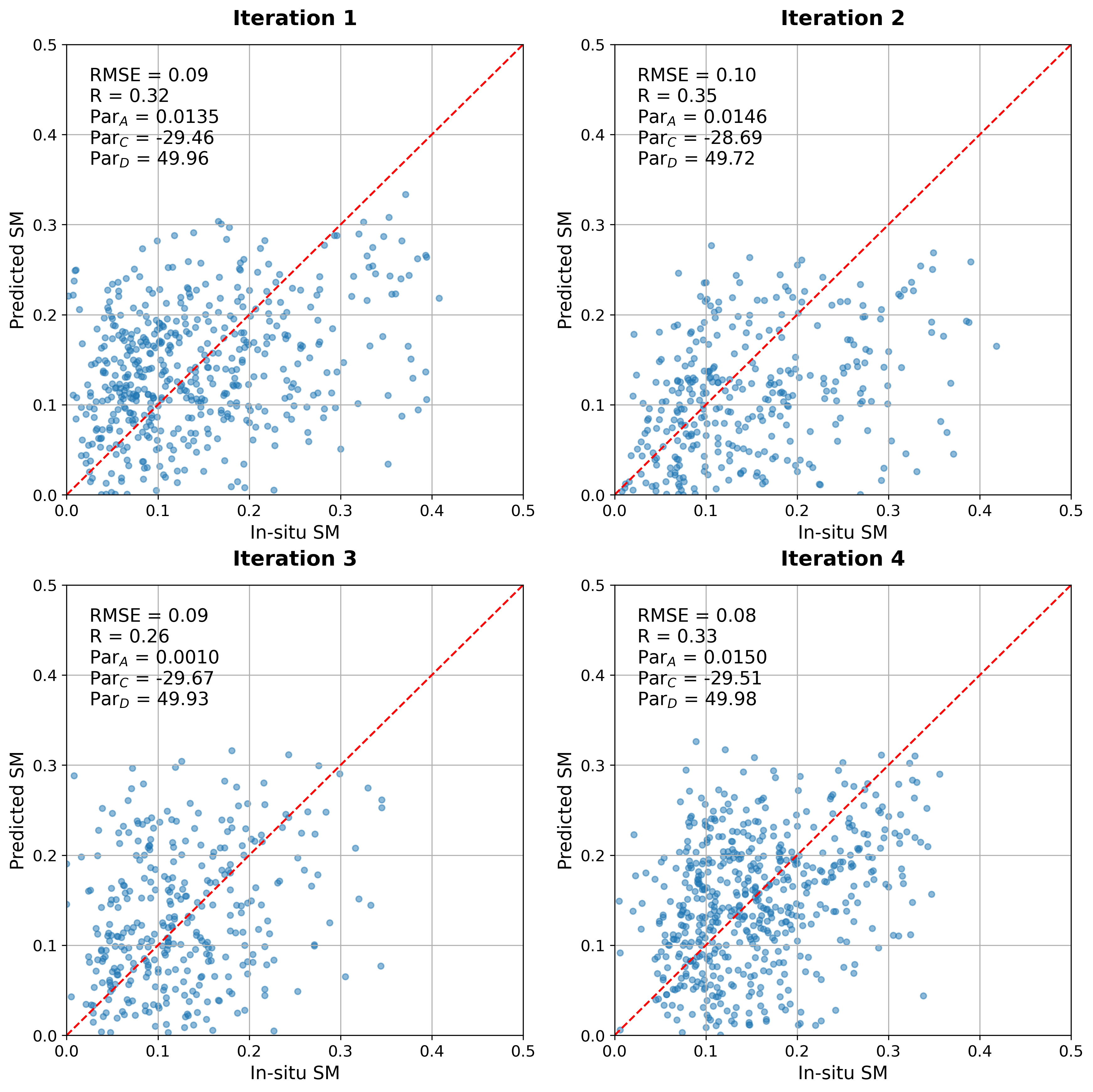}
    \caption{Scatterplots of WCM-predicted SM against in-situ SM in across the four-fold cross-validation. The unit is m$^3$/m$^3$. The red dashed line represents the 1:1 line.}
    \label{fig2}
\end{figure}

Fig. \ref{fig2} shows the scatterplots of WCM-predicted SM against in-situ SM across the four-fold cross-validation, with RMSE values ranging from 0.08 to 0.10 m$^3$/m$^3$ and correlation coefficients (R) between 0.26 and 0.34. All scatterplots exhibit substantial dispersion around the 1:1 line, with predicted SM values predominantly clustered in the intermediate range (0.15-0.25 m$^3$/m$^3$). This sub-optimal performance can be attributed to the limitations of the WCM framework in handling complex surface conditions. Specifically, the model's simplified linear regression (Eq. \ref{eq6}) fails to fully capture the non-linear relationships between soil backscatter and SM under varying surface characteristics. The assumption of a uniform vegetation layer and simplified scattering mechanisms in the WCM neglects important factors such as soil roughness variations, vegetation structural complexity, and the spatial heterogeneity of soil properties, leading to the observed dispersion and reduced sensitivity to moisture extremes.

\begin{figure}
    \centering
    \includegraphics[width=0.48\textwidth]{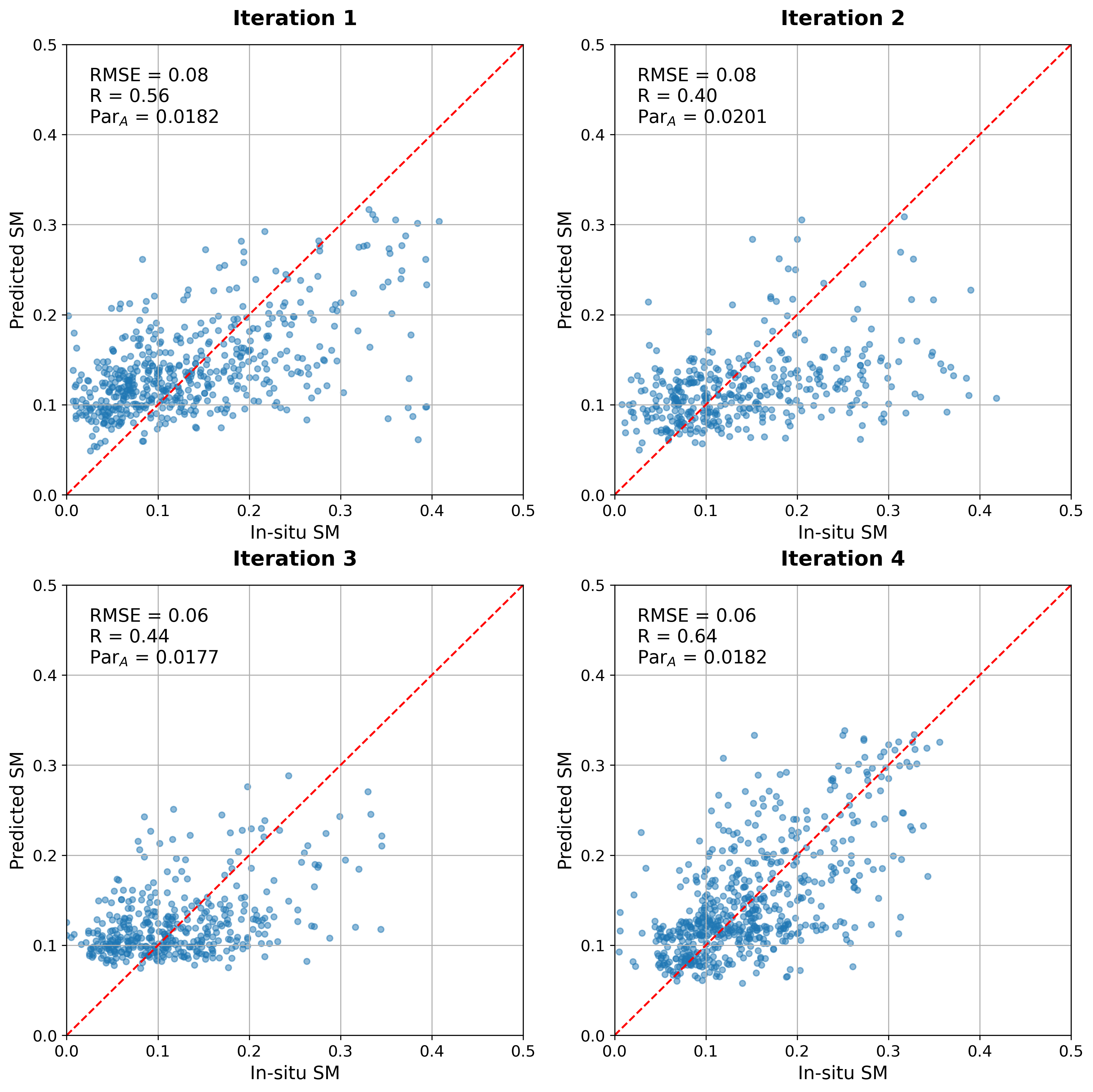}
    \caption{Scatterplots of SM predicted by the WCM knowledge-guided LSTM against in-situ SM across the four-fold cross-validation. The unit is m$^3$/m$^3$. The red dashed line represents the 1:1 line.}
    \label{fig3}
\end{figure}

Fig. \ref{fig3} shows the scatterplots of SM predicted by the WCM knowledge-guided LSTM against in-situ SM across the four-fold cross-validation. The results exhibit consistent RMSE values ranging from 0.06 to 0.08 m$^3$/m$^3$, indicating relatively robust estimation capabilities across different spatial contexts. The values of R vary from 0.40 to 0.64, with the iteration 4 showing the strongest relationship between predicted and observed values. The stability of vegetation backscattering parameter A (ranging from 0.0177 to 0.0201) across iterations suggests consistent characterisation of the vegetation-backscatter relationships. Notable is this approach's tendency to slightly underestimate SM at higher values ($>$0.3 m$^3$/m$^3$), particularly evident in the iterations 1-3, while the iteration 4 demonstrates improved performance across the full range of SM conditions. This validation reveals the spatial transferability of the proposed knowledge-guided approach while highlighting specific moisture ranges where further refinement may be beneficial.

\section{Conclusion}

This research presents a knowledge-guided deep learning approach that integrates WCM knowledge into a LSTM architecture for field-scale SM estimation. By incorporating semi-physical constraints through a dual-component loss function, the proposed approach demonstrates improved performance across spatially independent validation sets compared to the WCM approach, achieving RMSE values of 0.06-0.08 m$^3$/m$^3$ and correlation coefficients up to 0.64. The results highlight the potential of hybrid architectures in addressing the over-simplified parameterization of traditional SM retrieval methods, particularly in heterogeneous agricultural landscapes. Nonetheless, it is worthwhile exploring the potential of this approach across diverse spatial contexts in the future to assess its operational capability in SM monitoring.

\bibliographystyle{IEEEbib}
\bibliography{references}

\begin{thebibliography}{10}

\bibitem{seneviratne_investigating_2010}
Sonia~I Seneviratne, Thierry Corti, Edouard~L Davin, Martin Hirschi, Eric~B Jaeger, Irene Lehner, Boris Orlowsky, and Adriaan~J Teuling,
\newblock ``Investigating soil moisture–climate interactions in a changing climate: {A} review,''
\newblock {\em Earth-Science Reviews}, vol. 99, no. 3, pp. 125--161, 2010,
\newblock \url{https://doi.org/10.1016/j.earscirev.2010.02.004}.

\bibitem{babaeian_ground_2019}
Ebrahim Babaeian, Morteza Sadeghi, Scott~B Jones, Carsten Montzka, Harry Vereecken, and Markus Tuller,
\newblock ``Ground, proximal, and satellite remote sensing of soil moisture,''
\newblock {\em Reviews of Geophysics}, vol. 57, no. 2, pp. 530--616, 2019,
\newblock \url{https://doi.org/10.1029/2018rg000618}.

\bibitem{entekhabi_soil_2010}
D~Entekhabi, E~G Njoku, P~E~O' Neill, K~H Kellogg, W~T Crow, W~N Edelstein, J~K Entin, S~D Goodman, T~J Jackson, J~Johnson, J~Kimball, J~R Piepmeier, R~D Koster, N~Martin, K~C McDonald, M~Moghaddam, S~Moran, R~Reichle, J~C Shi, M~W Spencer, S~W Thurman, L~Tsang, and J~Van Zyl,
\newblock ``The {Soil} {Moisture} {Active} {Passive} ({SMAP}) {Mission},''
\newblock {\em Proceedings of the IEEE}, vol. 98, no. 5, pp. 704--716, 2010,
\newblock \url{https://doi.org/10.1109/jproc.2010.2043918}.

\bibitem{yu_continental_2021}
Yi~Yu, Luigi~J Renzullo, and Siyuan Tian,
\newblock ``Continental scale downscaling of {AWRA}-{L} analysed soil moisture using random forest regression,''
\newblock in {\em {MODSIM2021}, 24th {International} {Congress} on {Modelling} and {Simulation}}, Sydney, Australia, 2021, pp. 498--504,
\newblock \url{https://doi.org/10.36334/modsim.2021.j10.yu}.

\bibitem{yu_empirical_2024}
Yi~Yu, Brendan~P. Malone, and Luigi~J. Renzullo,
\newblock ``Empirical upscaling of point-scale soil moisture measurements for spatial evaluation of model simulations and satellite retrievals,''
\newblock in {\em 2024 {IEEE} {International} {Geoscience} and {Remote} {Sensing} {Symposium}}, Athens, Greece, July 2024, pp. 11496--11501,
\newblock \url{https://doi.org/10.1109/IGARSS53475.2024.10642763}.

\bibitem{peng_roadmap_2021}
Jian Peng, Clement Albergel, Anna Balenzano, Luca Brocca, Oliver Cartus, Michael~H Cosh, Wade~T Crow, Katarzyna Dabrowska-Zielinska, Simon Dadson, Malcolm W~J Davidson, Patricia de~Rosnay, Wouter Dorigo, Alexander Gruber, Stefan Hagemann, Martin Hirschi, Yann~H Kerr, Francesco Lovergine, Miguel~D Mahecha, Philip Marzahn, Francesco Mattia, Jan~Pawel Musial, Swantje Preuschmann, Rolf~H Reichle, Giuseppe Satalino, Martyn Silgram, Peter~M van Bodegom, Niko E~C Verhoest, Wolfgang Wagner, Jeffrey~P Walker, Urs Wegmüller, and Alexander Loew,
\newblock ``A roadmap for high-resolution satellite soil moisture applications – confronting product characteristics with user requirements,''
\newblock {\em Remote Sensing of Environment}, vol. 252, pp. 112162, 2021,
\newblock \url{https://doi.org/10.1016/j.rse.2020.112162}.

\bibitem{yu_spatial_2025}
Yi~Yu, Brendan~P. Malone, Luigi~J. Renzullo, Chad~A. Burton, Siyuan Tian, Ross~D. Searle, Thomas F.~A. Bishop, and Jeffrey~P. Walker,
\newblock ``Spatial soil moisture prediction from in-situ data upscaled to {Landsat} footprint: {Assessing} area of applicability of machine learning models,''
\newblock {\em IEEE Transactions on Geoscience and Remote Sensing}, 2025,
\newblock \url{https://doi.org/10.1109/TGRS.2025.3565818}.

\bibitem{fan_sentinel-1_2025}
Dong Fan, Tianjie Zhao, Xiaoguang Jiang, Almudena García-García, Toni Schmidt, Luis Samaniego, Sabine Attinger, Hua Wu, Yazhen Jiang, Jiancheng Shi, Lei Fan, Bo-Hui Tang, Wolfgang Wagner, Wouter Dorigo, Alexander Gruber, Francesco Mattia, Anna Balenzano, Luca Brocca, Thomas Jagdhuber, Jean-Pierre Wigneron, Carsten Montzka, and Jian Peng,
\newblock ``A {Sentinel}-1 {SAR}-based global 1-km resolution soil moisture data product: {Algorithm} and preliminary assessment,''
\newblock {\em Remote Sensing of Environment}, vol. 318, pp. 114579, 2025,
\newblock \url{https://doi.org/10.1016/j.rse.2024.114579}.

\bibitem{wagner_temporal_2008}
Wolfgang Wagner, Carsten Pathe, Marcela Doubkova, Daniel Sabel, Annett Bartsch, Stefan Hasenauer, Günter Blöschl, Klaus Scipal, José Martínez-Fernández, and Alexander Löw,
\newblock ``Temporal {Stability} of {Soil} {Moisture} and {Radar} {Backscatter} {Observed} by the {Advanced} {Synthetic} {Aperture} {Radar} ({ASAR}),''
\newblock {\em Sensors}, vol. 8, no. 2, pp. 1174--1197, 2008,
\newblock \url{https://doi.org/10.3390/s80201174}.

\bibitem{torres_gmes_2012}
Ramon Torres, Paul Snoeij, Dirk Geudtner, David Bibby, Malcolm Davidson, Evert Attema, Pierre Potin, BjÖrn Rommen, Nicolas Floury, Mike Brown, Ignacio~Navas Traver, Patrick Deghaye, Berthyl Duesmann, Betlem Rosich, Nuno Miranda, Claudio Bruno, Michelangelo L'Abbate, Renato Croci, Andrea Pietropaolo, Markus Huchler, and Friedhelm Rostan,
\newblock ``{GMES} {Sentinel}-1 mission,''
\newblock {\em Remote Sensing of Environment}, vol. 120, pp. 9--24, 2012,
\newblock \url{https://doi.org/10.1016/j.rse.2011.05.028}.

\bibitem{bauer-marschallinger_toward_2019}
B~Bauer-Marschallinger, V~Freeman, S~Cao, C~Paulik, S~Schaufler, T~Stachl, S~Modanesi, C~Massari, L~Ciabatta, L~Brocca, and W~Wagner,
\newblock ``Toward {Global} {Soil} {Moisture} {Monitoring} {With} {Sentinel}-1: {Harnessing} {Assets} and {Overcoming} {Obstacles},''
\newblock {\em IEEE Transactions on Geoscience and Remote Sensing}, vol. 57, no. 1, pp. 520--539, 2019,
\newblock \url{https://doi.org/10.1109/tgrs.2018.2858004}.

\bibitem{oh_empirical_1992}
Y~Oh, K~Sarabandi, and F~T Ulaby,
\newblock ``An empirical model and an inversion technique for radar scattering from bare soil surfaces,''
\newblock {\em IEEE Transactions on Geoscience and Remote Sensing}, vol. 30, no. 2, pp. 370--381, 1992,
\newblock \url{https://doi.org/10.1109/36.134086}.

\bibitem{ulaby_radar_1996}
Fawwaz~T Ulaby, Pascale~C Dubois, and Jakob van Zyl,
\newblock ``Radar mapping of surface soil moisture,''
\newblock {\em Journal of Hydrology}, vol. 184, no. 1, pp. 57--84, 1996,
\newblock \url{https://doi.org/10.1016/0022-1694(95)02968-0}.

\bibitem{attema_vegetation_1978}
E~P~W Attema and Fawwaz~T Ulaby,
\newblock ``Vegetation modeled as a water cloud,''
\newblock {\em Radio Science}, vol. 13, no. 2, pp. 357--364, Mar. 1978,
\newblock \url{https://doi.org/10.1029/rs013i002p00357}.

\bibitem{joseph_soil_2008}
A~T Joseph, R~van~der Velde, P~E O'Neill, R~H Lang, and T~Gish,
\newblock ``Soil {Moisture} {Retrieval} {During} a {Corn} {Growth} {Cycle} {Using} {L}-{Band} (1.6 {GHz}) {Radar} {Observations},''
\newblock {\em IEEE Transactions on Geoscience and Remote Sensing}, vol. 46, no. 8, pp. 2365--2374, 2008,
\newblock \url{https://doi.org/10.1109/tgrs.2008.917214}.

\bibitem{singh_piml-sm_2024}
A~Singh and K~Gaurav,
\newblock ``{PIML}-{SM}: {Physics}-{Informed} {Machine} {Learning} to {Estimate} {Surface} {Soil} {Moisture} {From} {Multisensor} {Satellite} {Images} by {Leveraging} {Swarm} {Intelligence},''
\newblock {\em IEEE Transactions on Geoscience and Remote Sensing}, vol. 62, pp. 1--13, 2024,
\newblock \url{https://doi.org/10.1109/TGRS.2024.3502618}.

\bibitem{karpatne_theory-guided_2017}
A~Karpatne, G~Atluri, J~H Faghmous, M~Steinbach, A~Banerjee, A~Ganguly, S~Shekhar, N~Samatova, and V~Kumar,
\newblock ``Theory-{Guided} {Data} {Science}: {A} {New} {Paradigm} for {Scientific} {Discovery} from {Data},''
\newblock {\em IEEE Transactions on Knowledge and Data Engineering}, vol. 29, no. 10, pp. 2318--2331, 2017,
\newblock \url{https://doi.org/10.1109/tkde.2017.2720168}.

\bibitem{zanaga_esa_2022}
Daniele Zanaga, Ruben Van De~Kerchove, Dirk Daems, Wanda De~Keersmaecker, Carsten Brockmann, Grit Kirches, Jan Wevers, Oliver Cartus, Maurizio Santoro, Steffen Fritz, Myroslava Lesiv, Martin Herold, Nandin-Erdene Tsendbazar, Panpan Xu, Fabrizio Ramoino, and Olivier Arino,
\newblock ``{ESA} {WorldCover} 10 m 2021 v200,'' 2022,
\newblock \url{https://doi.org/10.5281/ZENODO.7254221}.

\bibitem{smith_murrumbidgee_2012}
A~B Smith, J~P Walker, A~W Western, R~I Young, K~M Ellett, R~C Pipunic, R~B Grayson, L~Siriwardena, F~H~S Chiew, and H~Richter,
\newblock ``The {Murrumbidgee} soil moisture monitoring network data set,''
\newblock {\em Water Resources Research}, vol. 48, no. 7, 2012,
\newblock \url{https://doi.org/10.1029/2012wr011976}.

\bibitem{gorelick_google_2017}
Noel Gorelick, Matt Hancher, Mike Dixon, Simon Ilyushchenko, David Thau, and Rebecca Moore,
\newblock ``Google {Earth} {Engine}: {Planetary}-scale geospatial analysis for everyone,''
\newblock {\em Remote Sensing of Environment}, vol. 202, pp. 18--27, 2017,
\newblock \url{https://doi.org/10.1016/j.rse.2017.06.031}.

\bibitem{li_evaluation_2010}
F~Li, D~L~B Jupp, S~Reddy, L~Lymburner, N~Mueller, P~Tan, and A~Islam,
\newblock ``An {Evaluation} of the {Use} of {Atmospheric} and {BRDF} {Correction} to {Standardize} {Landsat} {Data},''
\newblock {\em IEEE Journal of Selected Topics in Applied Earth Observations and Remote Sensing}, vol. 3, no. 3, pp. 257--270, 2010,
\newblock \url{https://doi.org/10.1109/jstars.2010.2042281}.

\bibitem{schaaf_first_2002}
Crystal~B Schaaf, Feng Gao, Alan~H Strahler, Wolfgang Lucht, Xiaowen Li, Trevor Tsang, Nicholas~C Strugnell, Xiaoyang Zhang, Yufang Jin, Jan-Peter Muller, Philip Lewis, Michael Barnsley, Paul Hobson, Mathias Disney, Gareth Roberts, Michael Dunderdale, Christopher Doll, Robert~P d'Entremont, Baoxin Hu, Shunlin Liang, Jeffrey~L Privette, and David Roy,
\newblock ``First operational {BRDF}, albedo nadir reflectance products from {MODIS},''
\newblock {\em Remote Sensing of Environment}, vol. 83, no. 1, pp. 135--148, 2002,
\newblock \url{https://doi.org/10.1016/s0034-4257(02)00091-3}.

\bibitem{yu_generating_2023}
Yi~Yu, Luigi~J Renzullo, Tim~R McVicar, Brendan~P Malone, and Siyuan Tian,
\newblock ``Generating daily 100 m resolution land surface temperature estimates continentally using an unbiased spatiotemporal fusion approach,''
\newblock {\em Remote Sensing of Environment}, vol. 297, pp. 113784, 2023,
\newblock \url{https://doi.org/10.1016/j.rse.2023.113784}.

\bibitem{xing_retrieval_2025}
Zanpin Xing, Lin Zhao, Lei Fan, Gabrielle De~Lannoy, Xiaojing Bai, Xiangzhuo Liu, Jian Peng, Frédéric Frappart, Kun Yang, Xin Li, Zhilan Zhou, Xiaojun Li, Jiangyuan Zeng, Defu Zou, Erji Du, Chong Wang, Lingxiao Wang, Zhibin Li, and Jean-Pierre Wigneron,
\newblock ``Retrieval of 1 km surface soil moisture from {Sentinel}-1 over bare soil and grassland on the {Qinghai}-{Tibetan} {Plateau},''
\newblock {\em Remote Sensing of Environment}, vol. 318, pp. 114563, 2025,
\newblock \url{https://doi.org/10.1016/j.rse.2024.114563}.

\bibitem{hochreiter_long_1997}
S~Hochreiter and J~Schmidhuber,
\newblock ``Long {Short}-{Term} {Memory},''
\newblock {\em Neural Computation}, vol. 9, no. 8, pp. 1735--1780, 1997,
\newblock \url{https://doi.org/10.1162/neco.1997.9.8.1735}.

\end{thebibliography}

\end{document}